%% file: main.tex
\pdfoutput=1
\documentclass[10pt,twocolumn,letterpaper]{article}

\usepackage[pagenumbers]{cvpr} 

\input{preamble}

\definecolor{cvprblue}{rgb}{0.21,0.49,0.74}
\usepackage[pagebackref,breaklinks,colorlinks,allcolors=cvprblue]{hyperref}
\usepackage{multirow}
\usepackage{graphicx}
\usepackage{comment}
\usepackage{algorithm}
\usepackage{algpseudocode}
\usepackage{pifont}
\newcommand{\cmark}{\ding{51}} 
\newcommand{\xmark}{\ding{55}} 
\usepackage{subcaption,booktabs}
\usepackage[table]{xcolor}
\definecolor{lightcyan}{rgb}{0.88,1,1}
\definecolor{lavender}{RGB}{230,230,250}
\newcommand{\methodname}{\texttt{RFDM}\xspace}
\newcommand{\methodnameonefive}{\texttt{RFDM1.5}\xspace}
\newcommand{\methodnamethreefive}{\texttt{RFDM3.5}\xspace}
\makeatletter
\renewcommand{\thefootnote}{\arabic{footnote}}
\makeatother

\usepackage{amsmath}

\DeclareMathOperator*{\argmin}{arg\,min}

\usepackage{lipsum}

\newcommand\blfootnote[1]{%
  \begingroup
  \renewcommand\thefootnote{}\footnote{#1}%
  \addtocounter{footnote}{-1}%
  \endgroup
}

\usepackage{soul}
\usepackage[dvipsnames]{xcolor}

\def\paperID{} 
\def\confName{CVPR}
\def\confYear{2026}

\title{RFDM: Residual Flow Diffusion Model for Efficient Causal Video Editing \vspace{-0.8em}}

\author{%
Mohammadreza Salehi\textsuperscript{*}\textsuperscript{\dag},
~Mehdi Noroozi\textsuperscript{*},
~Luca Morreale, \\
Ruchika Chavhan, 
Malcolm Chadwick,
Alberto Gil Ramos, 
Abhinav Mehrotra \\
Samsung AI Center, Cambridge \vspace{-1.5em} \\
}

\begin{document}

\twocolumn[{%
\renewcommand\twocolumn[1][]{#1}%
     \maketitle
     \centering
     \includegraphics[width=\linewidth]{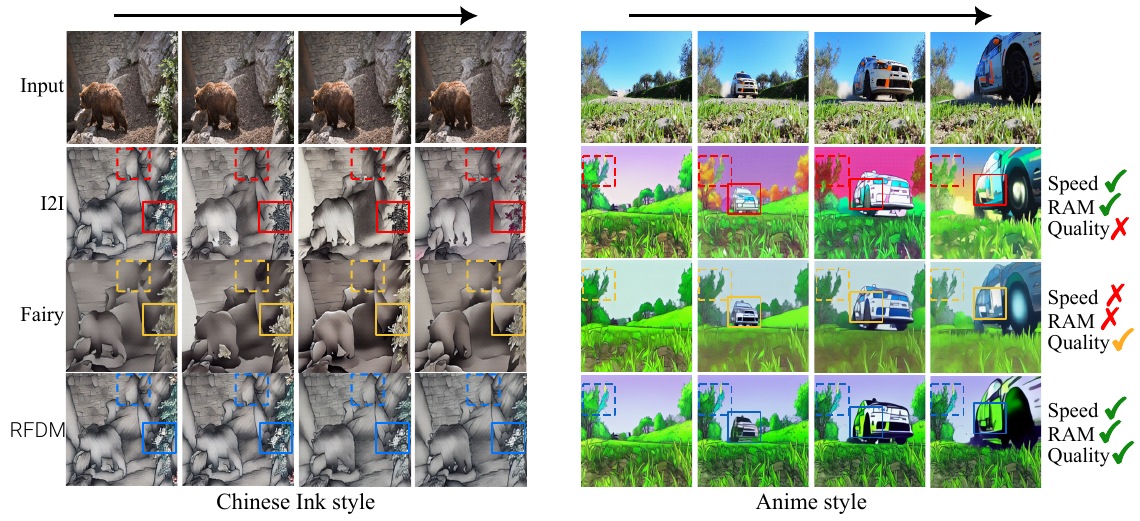}
     \vspace{-2em}
     \captionof{figure}{\textbf{Video style transfer.} Editing video style requires understanding of motion and style across frames. Naive Image-to-Image(I2I)~\cite{rombach2022high} models generate inconsistent video frames. Fairy~\cite{wu2024fairy} smoothens the result, trading computational cost to achieve lower jittering and inconsistency. \methodname results in the most consistent edit, while being computationally similar to the per-frame baseline (I2I).
     The video is taken from the DAVIS~\cite{pont20172017}.}
 \label{fig:teaser}
 }]
\blfootnote{*Equal contribution.}
\blfootnote{\dag Work done during the internship at Samsung.}


\input{sec/0_abstract}    
\input{sec/01_intro}
\input{sec/2_related_works}
\input{sec/3_method}

\input{sec/4_experiments}

\input{sec/6_conclusion}

\input{sec/X_suppl}

\clearpage
{
    \small
    \bibliographystyle{ieeenat_fullname}
    \bibliography{main}
}


\end{document}

%% file: preamble.tex
\usepackage[normalem]{ulem}

%% file: sec/0_abstract.tex
\begin{abstract}
Instructional video editing applies edits to an input video using only text prompts, enabling intuitive natural-language control. Despite the rapid progress, most methods still require fixed-length inputs and substantial compute. Meanwhile, autoregressive video generation enables efficient variable-length synthesis, yet remains under-explored for video editing. We introduce a \emph{causal}, \emph{efficient} video editing model that edits variable-length videos frame by frame. For efficiency, we start from a 2D image-to-image~(I2I) diffusion model and adapt it to video-to-video~(V2V) editing by conditioning the edit at time step $t$ on the model’s prediction at $t-1$. To leverage videos’ temporal redundancy, we propose a new I2I diffusion forward process formulation that encourages the model to predict the residual between the target output and the previous prediction. We call this \underline{R}esidual \underline{F}low \underline{D}iffusion \underline{M}odel~(\methodname), which focuses the denoising process on \emph{changes} between consecutive frames. Moreover, we propose a new benchmark that better ranks state-of-the-art methods by faithfulness for video editing tasks. Trained on paired video data for global/local style transfer and object removal, \methodname surpasses I2I-based methods and competes with fully spatiotemporal (3D) V2V models, while matching the compute of image models and scaling independently of input video length. More content can be found in \href{https://smsd75.github.io/RFDM_page/}{RFDM page}.
\end{abstract}


%% file: sec/01_intro.tex
\section{Introduction}

\label{sec:intro}

Instructional video editing aims to edit an input video using only a natural-language prompt (e.g., ``remove the object" or ``turn the object red"), eliminating extra conditioning signals such as object masks~\cite{xing2023survey, sun2024diffusion}. Recent methods have made strong progress by proposing different ways of coupling video generative models with text encoders or large language models~\cite{yu2025veggie, wang2025videodirector, shen2025qk, brooks2023instructpix2pix} or adapting an image-based model for video editing~\cite{wu2024fairy, li2024vidtome, qu2025tokenflow, kara2024rave}. However, all these approaches rely on non-causal temporal mechanisms that require fixed-length inputs and demand high computational cost, which is constraining in applications such as video streaming~\cite{li2025egoedit, chen2025streaming} or their deployment on resource-constrained devices such as cell phones~\cite{karnewar2025neodragon}.

Meanwhile, autoregressive video generation~\cite{yuan2025lumos,gu2025long,teng2025magi} has recently been recognized as a solution to variable-length input requirements and can be relatively efficient due to causal attention. However, such methods remain under-explored for video editing and still require further efficiency to operate in real time or on resource-constrained devices. We therefore propose a \emph{causal} and \emph{efficient} framework that edits videos autoregressively, frame by frame. For efficiency, we use 2D image-to-image~(I2I) diffusion models as the backbone and make them causal by (i) conditioning each prediction at time step $t$ on the previous-frame prediction as an input, which adds no extra compute, and (ii) Leveraging videos’ temporal redundancy, we propose a novel forward diffusion process that shifts the sampling-noise mean toward the previous prediction, turning current frame prediction into a residual prediction between consecutive frames. As shown in \autoref{fig:teaser} (I2I vs.\ \methodname), applying an I2I model independently to each frame yields inconsistent edits due to diffusion stochasticity or input frame variability; in contrast, our method produces consistent results on the same backbone, with no additional computational overhead.

Finally, we show that existing benchmarks that use metrics based on CLIP textual similarity between edited videos and prompts are insufficient to measure the faithfulness of edits, resulting in sub-optimal method rankings; we therefore introduce a new benchmark with new metrics and provide extensive qualitative and quantitative evaluations. We call our model \methodname~(\underline{Re}sidual \underline{F}low \underline{D}iffusion \underline{M}odel) and train it on the recently released Se\~norita~\cite{zi2025se} dataset. The evaluations on 3 benchmarks show that \methodname beats all models that are based on a similar 2D backbone and competes with models on a 3D backbone across three prompt categories: global style transfer, local style transfer, and object removal. In summary:
\begin{itemize}
    \item We propose a causal video editing model built on I2I backbones, solving the fixed-length input requirement while adding no overhead than an I2I model.\\
    \item We adapt the I2I backbone to videos by (i)~conditioning each time step on the previous prediction, and (ii)~introducing a novel diffusion forward process that encourages predicting inter-frame residuals rather than full frames.\\
    \item We present a new benchmark and accompanying metrics, and evaluate \methodname extensively on it and two additional benchmarks. \methodname achieves stronger faithfulness, competitive performance, and orders-of-magnitude lower latency compared to state-of-the-art.
\end{itemize}

%% file: sec/2_related_works.tex
\section{Related works}
\label{sec:related_works}
We categorize video editing models into text-guided and other modality-guided approaches.
Text-guided models edit videos solely based on textual prompts, enabling editing through natural language. In contrast, models in the latter group require additional inputs such as object masks~\cite{huang2025dive, kahatapitiya2024object}, optical flow~\cite{feng2024ccedit, hu2023videocontrolnet, tu2024motioneditor}, or audio cues~\cite{lee2023soundini}. In the following, we focus on text-guided methods, as they share the same setup as ours.

\noindent\textbf{Zero/few-shot methods.} These methods leverage pre-trained text-to-image~(T2I), image-to-image~(I2I), or text-to-video~(T2V) models and adapt them for video editing tasks in a zero-shot or few-shot manner~\cite{qu2025tokenflow, qi2023fatezero, cong2023flatten, ceylan2023pix2video, wu2023tune, wang2023zero, shen2025qk, huang2025videomage, wu2024fairy, li2024vidtome, wang2025videodirector, yang2025videograin}. T2I and I2I methods have similar computational costs and are substantially cheaper than T2V models. Zero-shot methods, such as VideoDirector~\cite{wang2025videodirector}, build upon pre-trained T2V models and steer the DDIM inversion process to achieve temporally consistent edits guided by text prompts. Similar ideas have been explored in FateZero~\cite{qi2023fatezero} and VideoGrain~\cite{yang2025videograin}, which adapt T2I backbones for video editing. Another line of zero-shot approaches enforces temporal coherence by aligning spatially similar features across frames, either through feature merging as in VidToMe~\cite{li2024vidtome} or spatio-temporal cross-attention as in Fairy~\cite{wu2024fairy}. In contrast, few-shot methods finetune the backbone to personalize edits using a small set of videos or images, as demonstrated in Tune-A-Video~\cite{wu2023tune}. While such methods are effective when the full video is accessible during inference, they cannot generate videos autoregressively—i.e., editing each frame based only on preceding frames—and either require fine-tuning the backbone for each new video or demand high computational costs due to heavy spatio-temporal attention across all frame tokens. \methodname, similar to Fairy and VidToMe, builds on a T2I backbone; however, it introduces no additional computational overhead beyond that of an image model, requires no per-video finetuning, and operates in an autoregressive manner.


\noindent\textbf{Training-based methods.} Existing approaches often depend on massive datasets and large video-based backbones~\cite{jiang2025vace, ye2025stylemaster, singer2024video}, which are difficult to access and reproduce. To address this, some methods~\cite{cheng2023consistent} train on synthetically generated data and adapt image-based models to videos by incorporating non-autoregressive spatio-temporal attention mechanisms, which increase computational cost with video length and perform less effectively on real-world videos. In contrast, \methodname\ is trained on the Señorita dataset~\cite{zi2025se}—the first large-scale, open-source video editing dataset constructed from real-world videos—and employs a novel autoregressive conditioning scheme that maintains the efficiency of image-based models without adding computational overhead.

%% file: sec/3_method.tex
\section{Method} \label{sec:method}

In this section, we discuss our proposed Residual Flow Diffusion Model (\methodname) for video editing. Our objective is to adapt an image-to-image~(I2I) model to an autoregressive video-to-video~(V2V) model. More precisely, given a series of input video frames denoted by $X = \{{x_t|t=0,\dots, T}\}$, and an instruction prompt $p$. We aim to train a diffusion model that edits the video auto-regressively frame-by-frame, producing a set of output frames denoted by $Y = \{{y_t^0|t=0,\dots, T}\}$.
To perform the edit consistently over the video, \methodname is conditioned on the previous prediction, $\hat{y}_{t-1}$.

In the following, we first present an overview of diffusion models for instruction-based image editing, then we discuss our proposed \methodname that adapts the I2I model to V2V. We denote the diffusion time-step by $s$ and the video temporal frames by $t$.

\subsection{Diffusion Models for Image Editing.} \label{sec:diff_img_edit}
Given a triplet $\{ x, p, y^0 \}$, the task of image editing involves modifying the input image, $x$, based on the instruction prompt, $p$, to closely match the target image $y^0$. Therefore, posing the task as a conditional image generation. Seminal work, Instruct-pix2pix~\cite{brooks2023instructpix2pix} poses it with the following diffusion forward process:

\begin{align}\label{eq:forward_image_edit}
q(y^s|y^0) & = \mathcal{N}(\alpha^s y^0 , (\sigma^s)^2 I); \quad s\in[0,1] \\ \nonumber
y^s & = \alpha^s y^0 + \sigma^s \epsilon; \quad \epsilon \sim \mathcal{N}(0,I) .
\end{align}

Where $\alpha^s, \sigma^s$ define the noise schedule, such that the log signal-to-noise ratio, $\lambda^s =  \log[\frac{\alpha^s}{\sigma^s}]$, decreases with $s$ monotonically.

The parameters of the denoising distribution, $\theta$, is learned via a denoising function through the following objective function\footnote{This objective function can be optimized with other parameterizations such as $\epsilon$, etc.~\cite{kingma2023understanding}. We use $x$ parameterization throughout this paper without losing generalizability.}:

\begin{equation}
\argmin_\theta ||\hat{\mathbf{y}}_\theta(y^s, x, p, \lambda_s) - y^0|| 
\end{equation}

Classifier Free Guidance (CFG) enables high-fidelity image sampling during inference with prompt adherence~\cite{ho2022classifier}, resulting in a procedure that combines three different calls of the conditional and unconditional denoising model.
%
\subsection{Residual Flow Diffusion Models}  \label{sec:rfdm}
The main challenge of converting an I2I to V2V lies in handling the temporal inconsistency across frames. Indeed, naively applying a I2I model independently per-frame generates inconsistent motion across frames. Our primary objective is to improve temporal consistency via conditioning on its own prediction. That is, for a given frame at time $t$, the model is conditioned on its own prediction at $t-1$, and is supervised during the training through consistent ground truth.

\vspace{-.4cm}
\paragraph{Frame prediction.}
A straightforward approach involves deploying a similar forward process as Eq.~\ref{eq:forward_image_edit} but conditioning only the denoising model. More precisely, the forward process for the frame at time $t$ has the following form:
\begin{align}\label{eq:forward_000}
q(y_t^s|y_t^0) & = \mathcal{N}(\alpha^s y_{t}^0, (\sigma^s)^2 I); \quad s\in[0,1] \\ \nonumber
y_t^s &= \alpha^s  y_t^0 + \sigma^s \epsilon; \qquad \epsilon(0,I) .
\end{align}

Additionally, the denoising process is conditioned on the previous prediction $\hat{y}_{t-1}$\footnote{We denote the past predictions with $\hat{y}$ because there are multiple options during the training, which we discuss later.}, resulting in the following objective function:

\begin{equation}\label{eq:loss_000}
\argmin_\theta ||\hat{\mathbf{y}}_\theta(y_t^s, \hat{y}_{t-1}, x_t, p, \lambda_s) - y_t^0|| ; \\ \nonumber 
\end{equation}
where $ t\in[0,T]$ and $\hat{y}_{-1} = 0$.
\vspace{-.4cm}
\paragraph{Residual Flow Prediction.}

Inspired by~\cite{yue2023resshift} on image inverse problems, we modify our forward process such that instead of generating the entire frame from pure noise, it rephrases the problem in the light residual generation between the target frames. In particular, we denote the temporal residual $m_t^0$ between the target and past prediction as  $m_t^0 = \hat{y}_{t-1} - y_t^0$. We use it to rephrase the forward process as follows:
\begin{align} \label{eq:forward}
q(y_t^s|y_t^0, \hat{y}_{t-1}) & = \mathcal{N}(\alpha^s  y_t^0 + \sigma^s \hat{y}_{t-1}, (\sigma^s)^2 I) \\ \nonumber
& = \mathcal{N}(\gamma^s y_{t}^0 + \sigma^s m_t^0, (\sigma^s)^2 I) \\ \nonumber .
\end{align}
where $\gamma^s = \sqrt{1-(\sigma^s)^2} + \sigma^s$.\footnote{Note that in flow models, $\gamma^s=1 \text{ } \forall s$. In the case of diffusion models, it is close to $1$ most of the time.} The forward process in Eq.~\ref{eq:forward} forms a Markov chain that transits the target at frame $t$ to the noisy version of the previous prediction through the temporal residual between them. This is equivalent to shifting the mean of the sampling noise in the diffusion process:

\begin{align}\label{eq:forward_001}
y_t^s & = \alpha^sy_t^0 + \sigma^s \hat{y}_{t-1} + \sigma^s \epsilon; \quad \epsilon \sim \mathcal{N}(0,I) \\ \nonumber
& = \alpha^sy_t^0 + \sigma^s \hat \epsilon; \qquad \qquad  \hat \epsilon \sim \mathcal{N}(\hat{y}_{t-1},I)
\end{align}

\begin{algorithm}[t]
\caption{\methodname training}\label{alg:train}
\begin{algorithmic}
\State \textbf{Input}: Input video clip $x_t = \{x_0, \dots, x_T\}$
\State \textbf{Input}: Output video clip $y_t = \{y^0_0, \dots, y^0_T\}$
\State \textbf{Input}: instruction prompt $p$
\State  \textbf{Input}: noise schedule $\alpha^s, \sigma^s, \lambda^s$ \Comment{for $s \in [0,1]$}
\State Input: initialization $\theta$
\Comment{from text-to-image}
\State Sample $K$ sorted indices $\{k_0,\dots, k_K \} \in[0,T]$
\State $\hat{y}_{-1} \leftarrow 0$ 
\State $\mathcal{L} \leftarrow 0$
\For{$i \in [0, \dots, K]$}
\State $t \leftarrow k_i$
\State  $s\sim \mathcal{U}(0,1)$
\State $\epsilon \sim \mathcal{N}(0,I)$ 
\State $y_t^s \leftarrow \alpha^sy^0_{t} + \sigma^s \hat{y}_{t-1} + \sigma^s \epsilon$
\State $\hat{y}_{t} \leftarrow \hat{\mathbf{y}}_{\theta}(y_t^s, x_t, \hat{y}_{t-1},p)$
\State $\mathcal{L} \mathrel{+}= \text{MSE}(\hat{y}_{t},{y}^0_{t}) $ \Comment{Apply the loss on all frames.}
\EndFor 
\State Update $\theta$ via backpacking through $\mathcal{L}$.
\end{algorithmic}
\end{algorithm}
\vspace{-6pt}

\paragraph{Denoising process.}
The denoising process estimates the conditional posterior distribution $p_\theta(y_t|y_t^s, \hat{y}_{t-1}, x_t, p)$ for which we train a denoising function $\hat{\mathbf{y}}_\theta(y_t^s, \hat{y}_{t-1}, x_t, p, \lambda_s)$ parametrized by a neural network in the following objective function:

\begin{equation}
\argmin_\theta ||\hat{\mathbf{y}}_\theta(y_t^s, \hat{y}_{t-1}, x_t, p, \lambda_s) - y_t^0|| .
\end{equation}

The advantage of Eq.~\ref{eq:forward_001} to ~\ref{eq:forward_000} lies in the supervision signal, which we elaborate in the following.

An efficient way for the network to optimize the denoising model is to split the pixels into two groups. I) The areas that are already edited in $\hat{y}_{t-1}$, such as background or a moving object that is present in both frames; the model needs to just shift them to $y_{t}$. II) The new areas, where the model needs to edit them based on $x_t$ and the instruction prompt. The forward process in Eq.~\ref{eq:forward_001} explicitly enforces this strategy, where the residual is explicitly embedded in the noisy input. 

\vspace{-.4cm}

\paragraph{Sampling.} During the sampling, we start with editing the first frame from the pure noise using the DDIM sampler. To edit the next frames, we start from a noise shifted by the previous frame prediction and repeat this process until the last frame. We use CFG to edit each frame as proposed in~\cite{brooks2023instructpix2pix}. 
\vspace{-.4cm}
\input{figures/training}
\paragraph{Handling the exposure bias.} Exposure bias refers to the discrepancy between the way an autoregressive model is used during the training and inference~\cite{huang2025self}, resulting in degraded video quality over time during the inference in the case of video generation.
Handling exposure bias translates to the choices for $\hat{y}_{t-1}$ in our notation, where, despite the inference time that we only have access to the model's inference prediction distribution, there are multiple choices to sample $\hat{y}_{t-1}$ during the training.  A straightforward choice is to use the ground truth frames, i.e. $y^0_{t-1}$, which is known as Teacher Forcing (TF)~\cite{williams1989learning}. However, as we ablate in the experiments section, the model trained with TF degrades quickly during inference. One reason could be the reliance on clean $y^0_{t-1}$ during the training, which suffers from a shift between the model's imperfect prediction distribution during the inference. 
To close the gap between the training and inference distribution of $\hat{y}_{t-1}$, we follow Diffusion Forcing (DF)~\cite{chen2024diffusion}. That is, we sample different random noise levels for different frames during the training to generate the noisy inputs, and use the denoising model prediction at frame $t-1$ to sample $\hat{y}_{t-1}$. As a result, the past observations are still clean, but they are sampled from the model's training distribution, which is closer to the model's inference distribution compared with the ground truth in TF. During the inference, we edit each frame in a fully sequential manner, and use the past frame's clean output to condition next frame editing. The training and inference processes are illustrated in detail in Algorithm~\ref{alg:train},  \autoref{fig:training} and Algorithm~\ref{alg:inference},~\autoref{fig:inference} respectively.

%% file: figures/training.tex
\begin{figure}[!t]
    \centering
    \includegraphics[width=1.\linewidth]{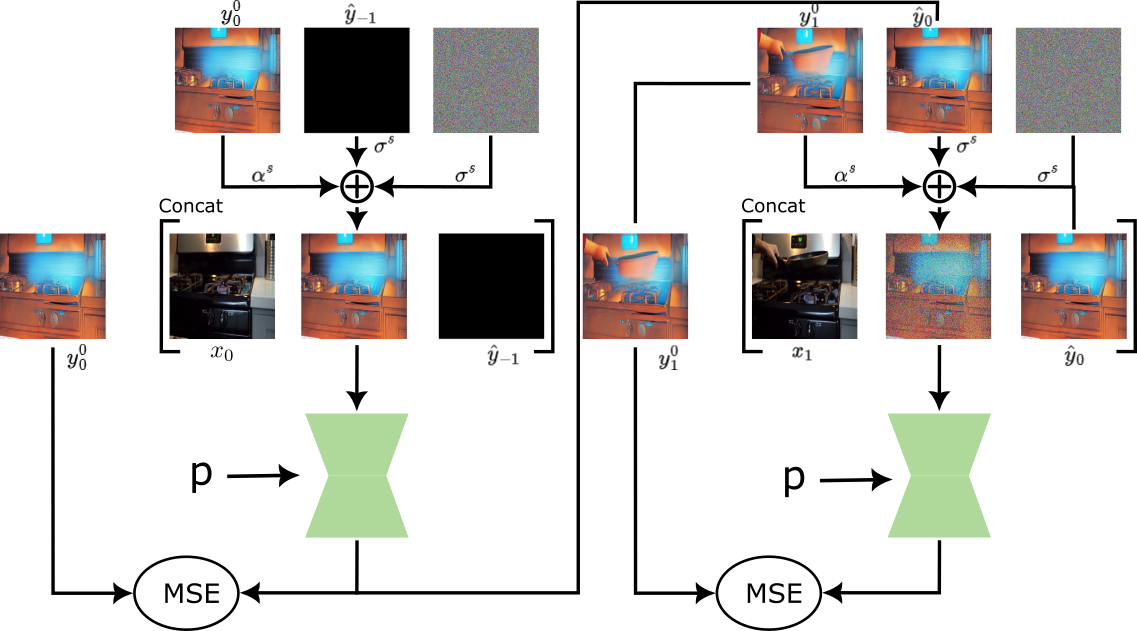}
    \caption{\footnotesize{\textbf{Training.}\label{fig:training}
    We obtain the noisy input($y_t^s$) at time frame $t$ by fusing current frame($x_t$), previous prediction($\hat{y}_{t-1}$), and noise through a noise scheduler via Eq.~\ref{eq:forward_001}. For a given video clip, we edit the first frame, where the noisy input is obtained by adding noise to the target; where the first target is $\hat{y}_{-1}=0$. Thereafter, the output on the previous frame is used as input ($\hat{y}_1$). The amount of noise, i.e.~$\alpha^s, \sigma^s$, is sampled independently for each frame. This figure shows a scenario where the amount of noise is low in the first frame, i.e.~$\alpha^s \rightarrow 1$, and high in the second frame, i.e.~$\alpha^s \rightarrow 0$. In the latter, where the amount of noise is high, $y_t^s$ is close to the noisy $\hat{y}_{t-1}$. In addition to $y_t^s$, the denoising model takes $x_t$ and $\hat{y}_{t-1}$ as extra inputs via concatenation and the instruction prompt via cross attention.}}
\end{figure}
\vspace{-.1cm}

%% file: sec/4_experiments.tex
\section{Experiments} \label{sec:exp}

\begin{algorithm}[!t]

\caption{\methodname inference}\label{alg:inference}
\begin{algorithmic}
\State \textbf{Input}: Input video clip $X = \{x_0, \dots, x_T\}$
\State \textbf{Input}: instruction prompt $p$
\State  \textbf{Input}: noise schedule $\alpha^s, \sigma^s, \lambda^s$ \Comment{for $s \in [0,1]$}
\State  \textbf{Input}: $\omega_x, \omega_{xp}$
 \Comment{cfg coefficients}
\State \textbf{Output}: Output video clip $\bar{Y} \leftarrow \{\}$
\State $\hat{y}_{-1} \leftarrow 0$ \Comment{previous frame perdition}
\For{$t \in {0, \dots, T}$}
\State $\epsilon  \sim \mathcal{N}(0,I)$ 
\State $\bar{y}_t^S \leftarrow  \hat{y}_{t-1} + \epsilon$ \Comment{initial noisy input}
\For{$s \in [S, \dots, 1]$}
\State $y^{\emptyset}_{pred} \leftarrow \hat{\mathbf{y}}_{\theta}(\bar{y}_t^s, \hat{y}_{t-1}, \emptyset, \emptyset, \alpha^s, \sigma^s)$  
\State $y^{x}_{pred} \leftarrow \hat{\mathbf{y}}_{\theta}(\bar{y}_t^s, \hat{y}_{t-1}, x_t, \emptyset, \alpha^s, \sigma^s)$
\State $y^{xp}_{pred} \leftarrow \hat{\mathbf{y}}_{\theta}(\bar{y}_t^s, \hat{y}_{t-1}, x_t , p, \alpha^s, \sigma^s)$  
\State $y_{pred} \leftarrow y_
{pred}^{\emptyset} + \omega_x(y^{x}_{pred} - y^{\emptyset}_{pred})$
\State $\qquad \qquad \qquad +\text{ }\omega_{xp}(y^{xp}_{pred} - y^{x}_{pred})$

\State $\bar{y}_t^{s-1}  = \text{DDIM}(y_{pred}, \bar{y}_t^s, \alpha^s, \sigma^s)$
\EndFor
\State $\hat{y}_{t-1} \leftarrow \bar{y}_t^0$
\State Add $\bar{y}_t^0$ to $\bar{Y}$
\EndFor

\State return $\bar{Y}$
\end{algorithmic}
\end{algorithm}

\subsection{Experimental setup}
We train \methodname on SD1.5~\cite{rombach2022high}~(\methodnameonefive) and SD3.5~(M)~\cite{esser2024scaling}~(\methodnamethreefive) as our backbones. For training, we utilize the recently introduced Señorita-2M~\cite{zi2025se} dataset, which contains two million paired videos covering five instructional video editing tasks: global style transfer, local style transfer, object removal, object addition, and object swap. As the dataset lacks predefined train, validation, and test splits, we introduce a fixed random split of 80\%, 15\%, and 5\%, respectively, for consistent training, evaluation, and test. We split the data between train ($80\%$), validation ($5\%$), and test ($15\%$). \methodname is trained on 8 A100 GPUs for 45k steps with a batch size of 8, gradient accumulation of 2, a learning rate of $1\text{e}^{-4}$  using the FusedAdam optimizer~\cite{kingma2014adam}.

\subsection{Evaluation benchmarks and metrics} \label{sec:eval_metrics}
\paragraph{Benchmarks.} We evaluate \methodname on three benchmarks: TGVE~\cite{cheng2023consistent}, TGVE+~\cite{singer2024video} and Señorita~\cite{zi2025se}. TGVE and TGVE+ capture prompt alignment based on detailed captions provided for each (input, output) video pair when ground-truth data is absent. This provides a useful proxy for prompt alignment and temporal coherence without ground truth; however, its reliance on text limits evaluation of two core aspects: \emph{temporal consistency}, \ie, preserving edits across frames, and \emph{faithfulness}, \ie, leaving unrelated regions unchanged. Therefore, as a third complementary benchmark, we leverage the ground-truth data in the Señorita dataset and introduce the \textit{Señorita Benchmark}, which enables direct comparison between edited outputs and reference videos on the test split of Señorita. This benchmark overcomes both limitations by enabling direct evaluation of temporal consistency and faithfulness to the input content. We report performance on the test set across three editing tasks: global style transfer, local style transfer, and object removal, which together provide a comprehensive assessment of each method’s versatility on real-world videos.
\input{figures/inference}
\vspace{-.2cm}
\input{figures/qualitative_results}
\input{Tables/main_results}
\paragraph{Metrics.} 
For TGVE and TGVE+, we report the metrics proposed for the benchmark, including CLIPFrame~\cite{wu2023cvpr}, PickScore~\cite{kirstain2023pick}, and $\text{ViCLIP}_{out}$~\cite{singer2024video}.  
For Señorita, we evaluate four complementary protocols that collectively assess the quality of generated videos from different perspectives: 
(i) Warping Error~\cite{lai2018learning} quantifies \textit{temporal consistency} by measuring optical-flow-based deformation between consecutive frames. 
(ii) ViDreamSim, which we introduce, extends DreamSim~\cite{fu2023dreamsim} to videos by comparing each output frame with its ground-truth counterpart, capturing \textit{faithfulness}. 
(iii) Directional Visual Similarity~(DVS)~\cite{noroozi2025guidance} computes the average similarity of direction vectors between the input–output and input–ground-truth frames using CLIP’s visual encoder, assessing whether the visual transformation aligns with the intended change in a text-free manner. 
(iv) Error Accumulation, which we also introduce in this work, measures the drift commonly observed~\cite{wang2025error, yin2025slow} in autoregressive models by quantifying how the distribution of later frames deviates from that of the first output frame. 
(v) MLLM-as-a-Judge prompts GPT-4o~\cite{hurst2024gpt} to rate how well the edited video fulfills the user instruction, providing a holistic score from 1 to 10.

We formally define ViDreamSim and Error Accumulation as follows:
\begin{equation} 
\textsc{ViDreamSim} = \frac{1}{T} \sum_{t=1}^{T} d(\bar{y}^0_t, y^{0}_t),
\label{eq:vidreamsim}
\end{equation}
\begin{equation}
\textsc{ErrAccu} = \frac{1}{T-1} \sum_{t=1}^{T} d(\bar{y}^0_t, \bar{y}^0_0),
\label{eq:erroraccum}
\end{equation}
where \(\bar{y}^0_t\) and \(y^{0}_t\) denote the output and ground-truth frames at time \(t\), respectively, \(d(\cdot,\cdot)\) represents a perceptual distance metric such as DreamSim, and \(T\) is the number of frames. Qualitative illustrations of error accumulation and temporal consistency, along with details of the MLLM-as-a-Judge protocol, are provided in the supplementary material.
\newline
\newline
\noindent\textbf{Baseline methods.} We compare \methodname against two categories of methods: those built upon fully spatio-temporal~(3D) V2V backbones, such as EVE~\cite{singer2024video}, and those based on I2I backbones~(2D), such as Fairy~\cite{wu2024fairy} and VidToMe~\cite{li2024vidtome}. To ensure a fair comparison, for the latter group, which employs the same backbone as ours, we report results using both their original SD1.5 model and our pretrained variant. We use the official repository for VidToMe~\cite{li2024vidtome} and reproduce Fairy~\cite{wu2024fairy} ourselves, as its implementation is not publicly available.

\subsection{Video editing results}

\noindent\textbf{TGVE, TGVE+ and Señorita benchmarks.} We compare \methodname against state-of-the-art nobaselines in both editing quality and computational efficiency in \autoref{tab:tgve_senorita_side_by_side}. On TGVE and TGVE+, \methodnamethreefive leads in 4 of 6 metrics across the two benchmarks and performs competitively with proprietary V2V models such as EVE, despite relying on a much more efficient I2I backbone. In efficiency, \methodnamethreefive matches Fairy's latency while using $\sim$$13\times$ less RAM, and reduces latency by $\sim$$4\times$ compared to other baselines. Even the smaller \methodnameonefive remains competitive with I2I state-of-tshe-art methods such as TokenFlow and RAVE on TGVE/TGVE+, while being $\sim$$12$--$16\times$ faster and requiring $\sim$$4$--$5\times$ less RAM. Notably, all \methodname variants achieve the highest CLIPFrame score, suggesting strong temporal consistency, driven by conditioning on $\hat{y}_{t-1}$ (the previous prediction) and predicting a residual flow that updates only changed regions. Compared to EVE, \methodnamethreefive attains slightly lower PickScore, which can be explained by EVE's substantially larger 3D backbone (4.4B vs.\ 2.5B parameters for SD3.5) and its much larger training set (34M vs.\ 2M videos in Se\~norita). Since EVE is closed-source, we cannot directly measure or report its computational cost.

On Señorita, we compare against prior methods, which use similar I2I backbones. As shown, both variants of \methodname consistently outperform all others, both with their original UNet and when re-trained with ours, across DVS, MLLM-Judge, and ViDreamSim, while maintaining comparable temporal consistency. The lower ViDreamSim and higher DVS scores show closer alignment with the ground truth, also confirmed by MLLM-Judge. The slightly lower temporal consistency compared to VidToMe stems from \methodname’s stronger faithfulness to the ground truth, whereas VidToMe produces smoother yet less accurate outputs. This underscores the advantage of the Señorita benchmark, as its use of ground-truth data enables more accurate and discriminative evaluation of model faithfulness and editing performance. More details in the supplementary material.
\vspace{-.4cm}
\paragraph{Qualitative results.} We qualitatively illustrate the results in \autoref{fig:qualitative_results} and \autoref{fig:different_styles}. As shown in \autoref{fig:qualitative_results}, \methodname produces cleaner edits in the object removal task, leaving fewer artifacts and consistently inpainting backgrounds. Likewise, for both local and global style transfer, it generates videos with the highest faithfulness and temporal consistency. For instance, Fairy fails to maintain faithfulness over time, turning both the liquid and the person pink, while VidToMe deviates from the ground truth from the very first frame. We illustrate the effect of applying multiple global style transfer prompts to the same video in \autoref{fig:different_styles}, demonstrating the method’s ability to generalize and produce diverse outputs from a single input. We provide an extensive qualitative evaluation in the supplementary material.

\input{Tables/main_ablations}

\input{figures/RFDM_style}

\subsection{Ablations}

\methodname's key parameters are ablated using our smaller variant, SD1.5 backbone, and the Señorita validation set for global style transfer. The number of frames used in the autoregressive loss is fixed to 3 in all ablations, except in the ablation studying this parameter. Our objective is to identify configurations that yield high temporal consistency, low error accumulation, and strong faithfulness to the ground truth. 
\newline
\newline
\textbf{Number of autoregressive frames.} In \autoref{table:autoregressive_frames}, we show that increasing the number of autoregressive frames in the loss leads to improved temporal consistency and reduced error accumulation. This is because training over longer temporal contexts exposes the model to its own prediction errors, yielding a stronger supervision signal. Therefore, we use 5 autoregressive frames as the default.  
\newline
\newline
\textbf{Input conditioning.} The previous prediction is already utilized in the sampling noise during the inference. We additionally provide the clean previous prediction, denoted by $\hat{y}_{t-1}$, as an extra input in each sampling step.  We ablate the impact of this extra input in \autoref{table:input_conditioning}, showing the importance of this extra input, which boosts the performance.
\newline
\newline
\textbf{Forcing approach.} \autoref{table:forcing_approach} ablates the effect of teacher forcing versus our adapted diffusion forcing, including an additional metric to assess output faithfulness. As shown, both methods achieve comparable temporal consistency, while teacher forcing yields slightly lower error accumulation. However, it diverges more from the ground-truth distribution, making it less suitable for inference. Hence, we adopt diffusion forcing in our main experiments.
\newline
\newline
\textbf{Key-frame update interval.} Error accumulation tends to increase when $\hat{y}_{t-1}$ is updated at every frame. To mitigate this, during inference we update it every $\Delta$ frames. As shown in \autoref{table:key-frame-update}, setting $\Delta=0$ means for a frame at time $t$, the first frame output is always used as $\hat{y}_{t-1}$, leading to very low accumulation error but high temporal inconsistency. Increasing $\Delta$ from 1 to 3 reduces the accumulation error by roughly $50\%$ (from $0.16$ to $0.07$) while causing only a minor drop in temporal consistency (around $10\%$), after which the benefits plateau, we set $\Delta=3$. 
\newline
\newline
\textbf{Grad propagation.} We ablate the impact of unrolling~\cite{pascanu2013difficulty} -- 
allowing or blocking the gradient flow -- in all autoregressive losses. As shown in \autoref{table:grad_propagation}, unrolling improves performance, as the model benefits from richer supervision through gradient backpropagation across all autoregressive stages. We use unrolling in our main results.
\newline
\newline
\textbf{Prediction task.} We compare the impact of residual-flow prediction and frame prediction formulations in \autoref{table:prediction_task}. As shown, residual-flow prediction results in lower error accumulation, which we attribute to its improved tracking capability that naturally emerges from the residual-flow formulation. To validate this, we selected 10 videos from DAVIS and segmented the first-frame objects using Qwen-Image-Edit~\cite{wu2025qwen}. We then conditioned both models, frame and residual-flow prediction, on Qwen-Image-Edit’s output with the same prompt and measured their tracking performance. As shown in \autoref{table:davis-tracking}, given the same first-frame output, the residual-flow model propagates edits more accurately across frames, demonstrating superior tracking ability.

%% file: figures/inference.tex
\begin{figure}[t!]
    \centering
    \includegraphics[width=1.\linewidth]{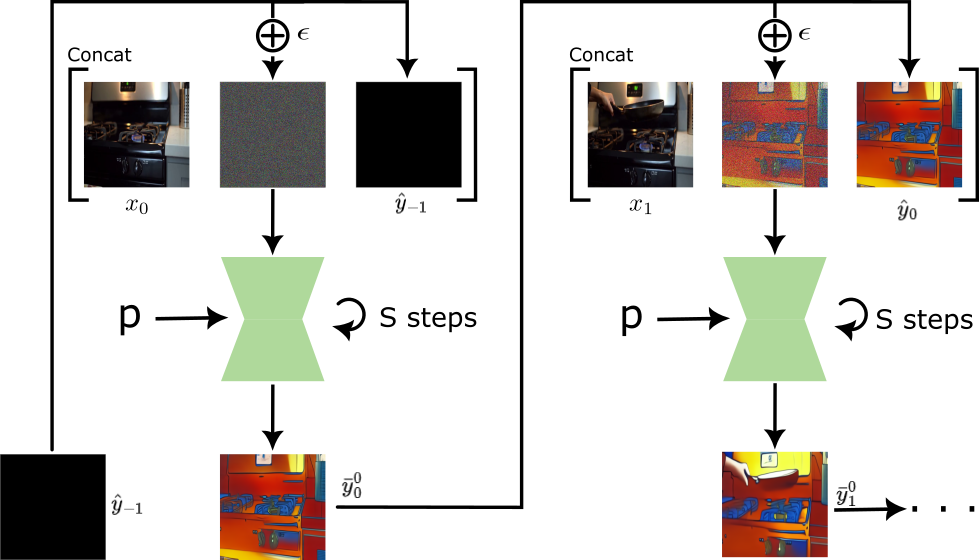}
    \caption{\footnotesize{\textbf{Inference illustration}.\label{fig:inference}
    We edit the frame at time $t$ by starting from the noisy previous frame~($\hat{y}_{t-1}$), and applying the denoising steps. The first frame, where $y_{-1}=0$, is edited in an I2I manner, starting from pure noise. In each denoising step, the current input frame($x_t$) and the clean previous prediction($\hat{y}_{t-1}$) are given to the model as extra input via concatenation to the noisy input. The instruction prompt is provided via cross-attention.}}  
\end{figure}

%% file: figures/qualitative_results.tex
\begin{figure*}[!htb]
    \centering
    \includegraphics[trim={4cm 0cm 5cm 0cm}, clip]{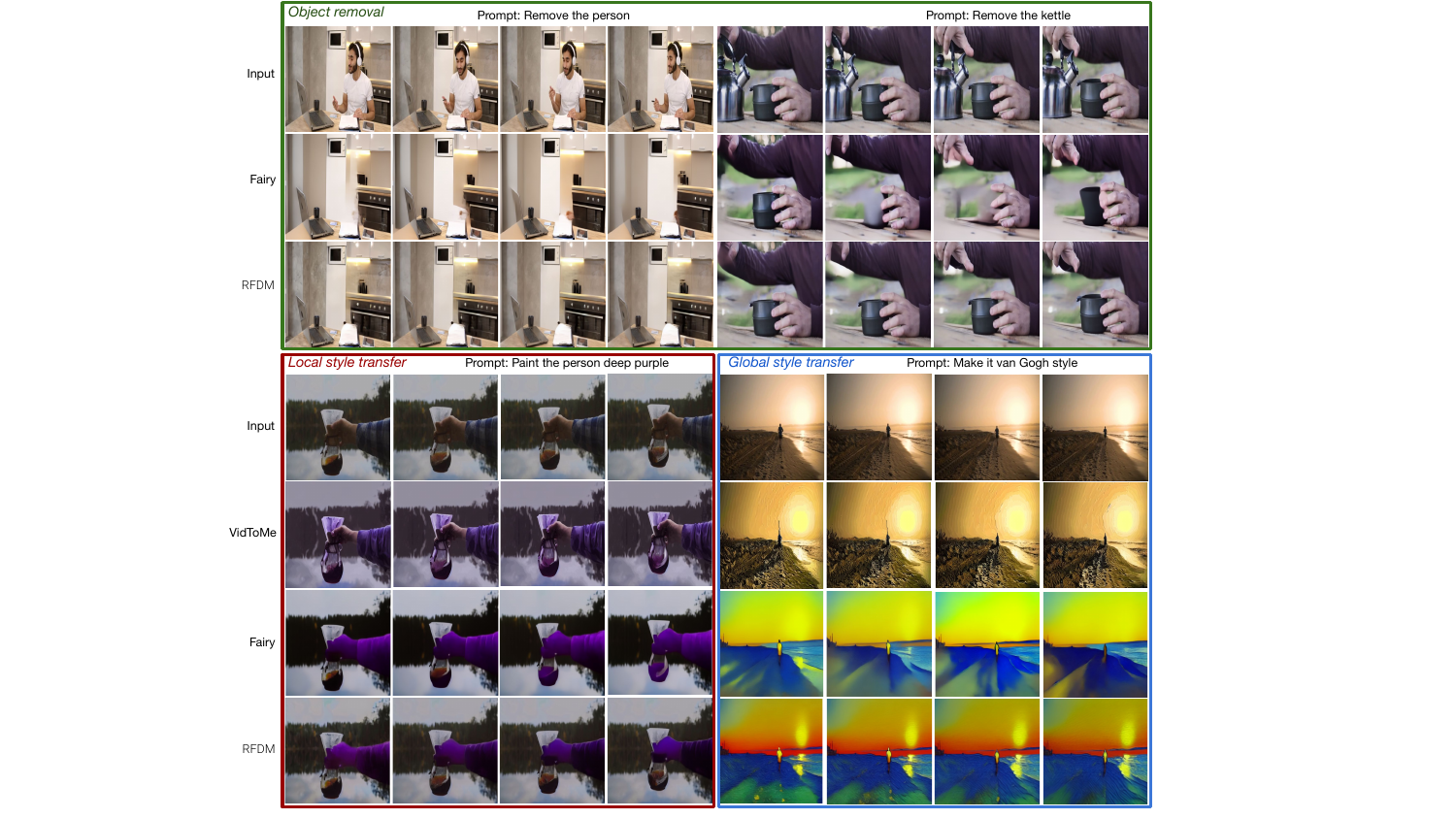}
    \vspace{-6pt}
    \caption{\textbf{Qualitative results.}\label{fig:qualitative_results} We compare \methodname with Fairy and VidToMe on Señorita benchmark across object removal (green), local style transfer (red), and global style transfer (blue). VidToMe is shown only for the style-transfer tasks as it is not designed for object removal. Compared to Fairy, \methodname yields more consistent and higher-fidelity outputs, especially in object removal where Fairy often leaves noticeable artifacts or alters irrelevant regions. VidToMe produces temporally consistent videos but deviates more from the ground truth than both \methodname and Fairy.}
    \vspace{-0.4cm}
\end{figure*}

%% file: Tables/main_results.tex
\begin{table*}[!htb]
\centering
\caption{\textbf{Video editing results on TGVE, TGVE+ and Señorita.} TGVE and TGVE+ results are taken from~\cite{singer2024video, zhang2025instructvedit} except for RAVE and TokenFlow, which we report. Señorita results are averaged across style transfer, local style transfer, and object removal tasks. TempCon denotes temporal consistency. 
*~marks methods using our pretrained UNet as the backbone. Latency and RAM are reported for generating 16 frames by us on an A100 GPU. Our models are highlighted in blue, and proprietary~(closed source and data) models are shown in \color{gray}{gray}.}
\label{tab:tgve_senorita_side_by_side}
\setlength{\tabcolsep}{2.2pt}
\renewcommand{\arraystretch}{1.10}

\resizebox{\textwidth}{!}{%
\begin{tabular}{l c|ccc|ccc|cccc|cc}
\toprule
& 
& \multicolumn{3}{c|}{\textbf{TGVE}} 
& \multicolumn{3}{c|}{\textbf{TGVE+}} 
& \multicolumn{4}{c|}{\textbf{Señorita test set}} 
& \multicolumn{2}{c}{\textbf{16-frame generation}} \\
\cmidrule(lr){3-5}\cmidrule(lr){6-8}\cmidrule(lr){9-12}\cmidrule(lr){13-14}
\textbf{Method} & \textbf{Causal}
& ViCLIP$_\text{out}\uparrow$ & PickScore$\uparrow$ & CLIPFrame$\uparrow$
& ViCLIP$_\text{out}\uparrow$ & PickScore$\uparrow$ & CLIPFrame$\uparrow$
& DVS$\uparrow$ & MLLM-Judge$\uparrow$ & ViDreamSim$\downarrow$ & TempCon$\downarrow$
& Lat.(s)$\downarrow$ & RAM(GB)$\downarrow$ \\
\midrule

Fairy~\cite{wu2024fairy} & \xmark
& 0.208 & 19.8 & 0.933
& 0.197 & 19.81 & 0.933
& 0.32 & 2.87 & 0.63 & 0.043
& \underline{13} & 77 \\

Fairy* & \xmark
& -- & -- & --
& -- & -- & --
& 0.40 & 3.98 & 0.29 & 0.042
& \underline{13} & 77 \\

AnyV2V~\cite{ku2024anyv2v} & \xmark
& 0.230 & 19.70 & 0.919
& 0.227 & 19.80 & 0.919
& -- & -- & -- & --
& 100 & 15 \\

STDF~\cite{yatim2024space} & \xmark
& 0.226 & 20.40 & 0.933
& 0.227 & \textbf{20.60} & 0.933
& -- & -- & -- & --
& 300 & 20 \\

TokenFlow~\cite{qu2025tokenflow} & \xmark
& 0.257 & 20.58 & 0.943
& 0.231 & 19.88 & 0.943
& 0.29 & 3.23 & 0.48 & 0.010
& 128 & 11 \\

RAVE~\cite{kara2024rave} & \xmark
& 0.254 & 20.35 & 0.936
& \textbf{0.250} & 20.03 & 0.932
& 0.34 & 3.54 & 0.42 & 0.017
& 92 & 9 \\

InsV2V~\cite{cheng2023consistent} & \xmark
& \underline{0.262} & \textbf{20.76} & 0.911
& 0.236 & 20.37 & 0.925
& -- & -- & -- & --
& 53 & 8 \\

VidToMe~\cite{li2024vidtome} & \xmark
& -- & -- & --
& -- & -- & --
& 0.32 & 3.10 & 0.46 & \textbf{0.007}
& 86 & 9 \\

VidToMe* & \xmark
& -- & -- & --
& -- & -- & --
& 0.37 & 1.77 & 0.59 & 0.014
& 86 & 9 \\

\color{gray}EVE~\cite{singer2024video} & \color{gray}\xmark
& \color{gray}0.262 & \color{gray}20.76 & \color{gray}0.922
& \color{gray}0.251 & \color{gray}20.88 & \color{gray}0.926
& \color{gray}-- & \color{gray}-- & \color{gray}-- & \color{gray}--
& \color{gray}-- & \color{gray}-- \\

\rowcolor{lightcyan}\methodnameonefive & \cmark
& 0.260 & 20.29 & \underline{0.949}
& \underline{0.236} & 19.33 & \underline{0.964}
& \underline{0.43} & \underline{6.60} & \underline{0.23} & 0.010
& \textbf{8} & \underline{2} \\

\rowcolor{lightcyan}\methodnamethreefive & \cmark
& \textbf{0.284} & \underline{20.66} & \textbf{0.965}
& \textbf{0.250} & \underline{20.52} & \textbf{0.970}
& \textbf{0.48} & \textbf{7.37} & \textbf{0.20} & \underline{0.009}
& \underline{13} & \textbf{6} \\

\bottomrule
\end{tabular}
} 
\vspace{-0.3cm}
\end{table*}

%% file: Tables/main_ablations.tex
\begin{table*}[!htb]
\caption{\textbf{Ablating the key parameters of, \methodname.} We report temporal consistency and error accumulation on the Señorita global style transfer validation set. $\uparrow$ and $\downarrow$ indicate that higher or lower values are better, respectively. The blue row, is our default setup.}
\label{tab:full_abl}
\setlength{\tabcolsep}{3pt}
\centering

\begin{subtable}[t]{0.24\textwidth}
    \centering
    \input{Tables/ar_abl}
\end{subtable}
\hfill
\begin{subtable}[t]{0.24\textwidth}
    \centering
    \input{Tables/conditioning_abl}
\end{subtable}
\hfill
\begin{subtable}[t]{0.4\textwidth}
    \centering
    \input{Tables/forcing_abl}
\end{subtable}

\vspace{3mm}

\begin{subtable}[t]{0.2\textwidth}
    \centering
    \input{Tables/chunk_abl}
\end{subtable}
\hfill
\begin{subtable}[t]{0.22\textwidth}
    \centering
    \input{Tables/unrolling_abl}
\end{subtable}
\hfill
\begin{subtable}[t]{0.22\textwidth}
    \centering
    \input{Tables/noise_abl}
\end{subtable}
\hfill
\begin{subtable}[t]{0.22\textwidth}
    \centering
    \caption{DAVIS Tracking}
    \input{Tables/davis_tracking} 
    \label{table:davis-tracking}
\end{subtable}


\vspace{-0.5cm}
\end{table*}

%% file: Tables/ar_abl.tex
\centering
\setlength{\tabcolsep}{3pt}
\caption{Autoregressive frames}
\footnotesize
{\begin{tabular}{ccc}
    \toprule
    \textsc{Num.} & \textsc{TempCon}$\downarrow$ & \textsc{ErrAccu}$\downarrow$ \\
    \midrule
    0   & 0.068 & 0.21 \\
    1 & 0.013 & 0.12 \\
    3 & 0.009 & 0.07 \\
   \rowcolor{lightcyan} 5 & 0.007 & 0.07\\
    \bottomrule
\end{tabular}
\label{table:autoregressive_frames}}

%% file: Tables/conditioning_abl.tex
\centering
\setlength{\tabcolsep}{3pt}
\caption{Input conditioning}
\footnotesize
{\begin{tabular}{lcc}
    \toprule
    \textsc{Inp.} & \textsc{TempCon}$\downarrow$ & \textsc{ErrAccu}$\downarrow$  \\
    \midrule
    $x$ & 0.027 & 0.14 \\
    \rowcolor{lightcyan} $x$,$\hat{y}_{t-1}$ & 0.009 & 0.07 \\
    \bottomrule
\end{tabular}
\label{table:input_conditioning}}

%% file: Tables/forcing_abl.tex
\centering
\setlength{\tabcolsep}{3pt}
\caption{Forcing approach}
\footnotesize
{\begin{tabular}{lccc}
    \toprule
    \textsc{Method} & \textsc{TempCon}$\downarrow$ & \textsc{ErrAccu}$\downarrow$ & \textsc{ViDreamSim}$\downarrow$ \\
    \midrule
    Teacher & 0.009 & 0.06 & 0.38\\
   \rowcolor{lightcyan} Diffusion & 0.009 & 0.07 & 0.35\\
    \bottomrule
\end{tabular}
\label{table:forcing_approach}}

%% file: Tables/chunk_abl.tex
\centering
\setlength{\tabcolsep}{1pt}
\caption{Key-frame update interval}
\footnotesize
{\begin{tabular}{ccc}
    \toprule
    $\Delta$Frames & \textsc{TempCon}$\downarrow$ & \textsc{ErrAccu}$\downarrow$  \\
    \midrule
    0 & 0.018 & 0.04 \\
    1 & 0.008 & 0.16 \\
   \rowcolor{lightcyan}  3 & 0.009 & 0.07 \\
    5 & 0.010 & 0.07 \\
    \bottomrule
\end{tabular}
\label{table:key-frame-update}}

%% file: Tables/unrolling_abl.tex
\centering
\setlength{\tabcolsep}{1pt}
\caption{Grad propagation}
\footnotesize
{\begin{tabular}{lcc}
    \toprule
    \textsc{Unrolling} & \textsc{TempCon}$\downarrow$ & \textsc{ErrAccu}$\downarrow$  \\
    \midrule
    No & 0.011 & 0.10\\
   \rowcolor{lightcyan} Yes & 0.009 & 0.07 \\
    \bottomrule
\end{tabular}
\label{table:grad_propagation}}

%% file: Tables/noise_abl.tex
\centering
\setlength{\tabcolsep}{1pt}
\caption{Prediction task formulation}
\footnotesize
{\begin{tabular}{lcc}
    \toprule
    \textsc{Noise} & \textsc{TempCon}$\downarrow$ & \textsc{ErrAccu}$\downarrow$  \\
    \midrule
    Frame & 0.009 & 0.09 \\
   \rowcolor{lightcyan} Residual-Flow & 0.009 & 0.07 \\
    \bottomrule
\end{tabular}
\label{table:prediction_task}}

%% file: Tables/davis_tracking.tex


\setlength{\tabcolsep}{1pt}
\footnotesize
\begin{tabular}{lccc}
\toprule
Prediction & J$\uparrow$ & F$\uparrow$ & J\&F$\uparrow$ \\
\midrule
Frame    & 33.3   & 24.9 & 29.1    \\
\rowcolor{lightcyan}Residual-Flow  & 50.6   & 36.4 & 43.6  \\
\bottomrule
\end{tabular}

%% file: figures/RFDM_style.tex
\begin{figure}[!htb]
    \centering
    \includegraphics[width=1.\linewidth, trim={1.5cm 5cm 13cm 0cm}, clip]{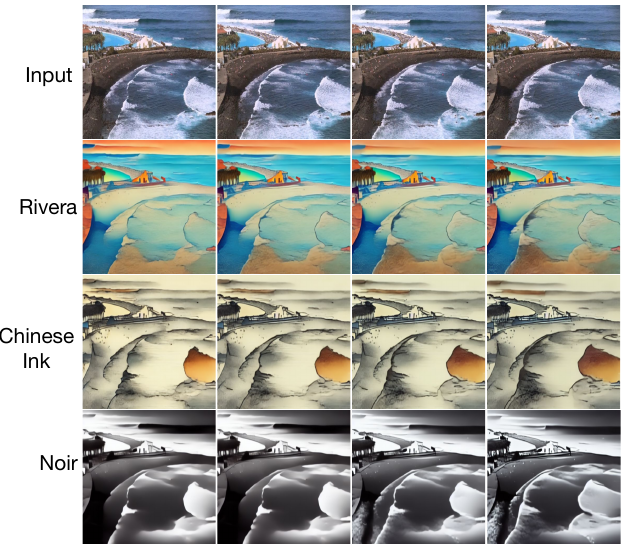}
    \vspace{-15pt}
    \caption{\textbf{Different styles generated by \methodname on Señorita.} \methodname produces consistent and faithful outputs across a wide range of styles and input videos.}
    \vspace{-10pt}
    \label{fig:different_styles}
\end{figure}

%% file: sec/6_conclusion.tex

\section{Conclusion and Discussion}

In this paper, we introduced \methodname, an autoregressive video editing model that adapts an image-to-image backbone through a novel autoregressive conditioning based on residual-flow prediction. Trained on large-scale real-world videos from the Señorita dataset, we extensively ablated across three key aspects: temporal consistency, error accumulation, and faithfulness, and evaluated on TGVE, TGVE+, and the proposed Señorita benchmark. The results show that causal I2I models like \methodname achieve strong performance and efficiency, pointing to a promising direction for scalable video editing. \textit{Limitations:} Despite an excellent trade-off between the performance and computational efficiency, \methodname's drawback lies in its short temporal memory, required in motion-constrained edits such as changing the action of a video. A plausible solution could be deploying a KV caching mechanism, which we leave to future work.

%% file: sec/X_suppl.tex
\clearpage
\setcounter{page}{1}
\maketitlesupplementary

\section{Experiments}
\label{sec:App_experiments}

\input{Tables/main_tables_full}

\subsection{Señorita Benchmark Details}

\autoref{tab:senorita_tasks_full} reports the full quantitative results on the Señorita benchmark for our method, along with Fairy~\cite{wu2024fairy} and VidToMe~\cite{li2024vidtome}, including per-task breakdown. We report the average scores, whereas here we show the complete tables for global style transfer, local style transfer, and object removal.

Overall, both variants of \methodname consistently improve over prior work across the majority of metrics, while maintaining orders of magnitude lower computational cost as shown in the paper Table 1. In particular, \methodnamethreefive attains higher or comparable scores in both temporal consistency and faithfulness metrics, which aligns with our design choices of residual flow prediction and autoregressive conditioning. For completeness, we also include the corresponding qualitative videos and error plots in the accompanying HTML visualization page (project page linked in the supplementary material).

\section{More qualitative results}

In addition to the figures shown in the main paper, we provide a larger set of qualitative examples (149 videos) in the html page provided with this supplementary material. We kindly encourage the reviewer to open the page for more comprehensive qualitative comparison.

\subsection{Failure cases}

We summarize common failure cases of \methodname, with more detailed explanations and video examples provided in the HTML page attached to the supplementary material. We encourage the reviewer to refer to the HTML visualization page for detailed explanations and examples.

\subsection{Better than ground-truth}

We also observe several cases where \methodname produces results that appear more natural than the provided ground truth. Since the Señorita dataset’s ground truth is generated through a multi-stage pipeline combining segmentation, inpainting, and tracking modules, it can occasionally contain artifacts, flickering, or imperfect boundaries. Benefiting from the strong spatial understanding of stable diffusion models and the temporal stability introduced by our proposed framework, \methodname often yields smoother and more coherent outputs that correct such imperfections. We encourage the reviewer to refer to the HTML visualization page in the supplementary material for side-by-side video comparisons.

\subsection{Visualization of ablation studies}
\label{sec:visualization_ablation}

We visualize what is measured in our ablation studies through \textbf{error accumulation} and \textbf{temporal consistency} in the attached HTML page. As shown, when \(\Delta = 0\), error accumulation remains low but temporal consistency degrades over time. 
At \(\Delta = 1\), temporal consistency improves, though accumulated error increases. 
The balance is achieved at \(\Delta = 3\), where both metrics remain stable. 
We encourage the reviewer to refer to the HTML visualization page in the supplementary material for corresponding video examples and plots.

\section{MLLM-as-a-Judge}
\label{sec:mllm_as_judge}

To obtain a human-aligned and instruction-aware evaluation of video editing quality, we employ a multimodal LLM-based scoring mechanism, referred to as \emph{MLLM-as-a-Judge}. This evaluator uses GPT-4o~\cite{hurst2024gpt} to assess how well an edited video satisfies the user instruction while preserving the structure and semantics of the original content.

\paragraph{Evaluation protocol.}
For each video, we consider the input sequence $x$, the predicted edited video $\bar{y}^0_t$, and the ground-truth edit $y^0_t$. We uniformly sample $K$ frames from each sequence to form triplets
\[
\left(x_{k},\, \bar{y}^0_{k},\, y^0_{k}\right), \qquad k = 1,\dots,K.
\]
Each triplet is processed independently to obtain frame-level judgments.

\paragraph{Prompting the MLLM.}
We provide the multimodal LLM with three images—the original frame, the edited frame produced by the method, and the ground-truth edited frame—along with the editing instruction. The model is asked to judge, in a comparative manner, how well the candidate edit adheres to the instruction while maintaining the spatial layout and identity of the original scene.

We use the following fixed prompt template:
\begin{quote}
\small
\texttt{You are an expert judge for video editing quality. You are given: (1) the original frame, (2) a candidate edited frame, and (3) the ground-truth edited frame, along with the editing instruction: ``\{$p$\}''. Rate how well the candidate frame follows the instruction while preserving the content, structure, and visual coherence of the original frame. Return a single score from 1 to 10, where 1 means ``much worse than the ground truth'' and 10 means ``better than the ground truth''. Output only the number.}
\end{quote}

\paragraph{Score aggregation.}
For each frame $k$, the MLLM outputs a scalar score $s_{k} \in \{1,\dots,10\}$.  
The video-level score is the average across sampled frames:
\[
\mathrm{MLLM\text{-}Judge}(x, \bar{y}^0, y^0)
= \frac{1}{K} \sum_{k=1}^K s_{k}.
\]
Finally, the dataset-level score for a method is obtained by averaging over all videos. A higher score indicates that the edited frames are judged to:
\begin{itemize}
    \item more faithfully follow the user-provided editing instruction,
    \item better preserve the identity, geometry, and temporal cues of the input,
    \item exhibit stronger perceptual coherence compared to the ground-truth edit.
\end{itemize}
Because the evaluation explicitly incorporates instruction text and visual context, it captures aspects of editing quality that traditional pixel- or feature-based metrics fail to measure.

The same protocol is used for all video-editing tasks considered in this work, including global style transfer, local style transfer, and object removal.

%% file: Tables/main_tables_full.tex

\begin{table*}[htb]
  \centering
  \caption{Señorita results are averaged across style transfer, local style transfer, and object removal tasks. TempCon denotes temporal consistency. 
*~marks methods using our pretrained SD1.5 UNet as the backbone.}
  \label{tab:senorita_tasks_full}
  \resizebox{\textwidth}{!}{
  \begin{tabular}{llcccc}
    \toprule
    \textbf{Task} & \textbf{Method} & DVS$\uparrow$ & \textsc{MLLM-Judge}$\uparrow$ & \textsc{TempCon}$\downarrow$ & \textsc{ViDreamSim}$\downarrow$ \\
    \midrule
    \multirow{6}{*}{Style transfer}
      & Fairy~\cite{wu2024fairy}      & 0.49 & 2.5 & 0.045 & 0.77 \\
      & Fairy*     & 0.56 & 3.7 & 0.048 & 0.50 \\
      & VidToMe~\cite{li2024vidtome}    & 0.51 & 3.5 & \textbf{0.007} & 0.60 \\
      & VidToMe*   & 0.57 & 1.5 & 0.015 & 0.51 \\
    \rowcolor{lightcyan}  & \methodnameonefive       & \underline{0.59} & \underline{6.0} & \textbf{0.007} & \underline{0.43} \\
    \rowcolor{lightcyan} & \methodnamethreefive       & \textbf{0.63} & \textbf{7.5} & \underline{0.008} & \textbf{0.36} \\
    \midrule
    \multirow{6}{*}{Local editing}
      & Fairy~\cite{wu2024fairy}      &  0.21 & 3.0 & 0.037 & 0.65 \\
      & Fairy*     & 0.29 & 4.0 & 0.037 & 0.17 \\
      & VidToMe~\cite{li2024vidtome}    & 0.22 & 2.8 & \textbf{0.007} & 0.35 \\
      & VidToMe*   & 0.26 & 1.5 & 0.015 & 0.61 \\
     \rowcolor{lightcyan} & \methodnameonefive       & \underline{0.34} & \underline{6.3} & 0.012 & \underline{0.13} \\
      \rowcolor{lightcyan} & \methodnamethreefive       & \textbf{0.40} & \textbf{6.8} & \underline{0.010} & \textbf{0.12} \\
    \midrule
    \multirow{6}{*}{Object removal}
      & Fairy~\cite{wu2024fairy}      & 0.26 & 3.1  & 0.046 & 0.46 \\
      & Fairy*     & 0.35 & 4.2 & 0.042 & 0.19\\
      & VidToMe~\cite{li2024vidtome}    & 0.23 & 3.0 & \textbf{0.007} & 0.43\\
      & VidToMe*   & 0.29 & 2.3 & 0.012 & 0.65\\
     \rowcolor{lightcyan} & \methodnameonefive       & \underline{0.37} & \underline{7.5} & 0.011 & \underline{0.12}\\
    \rowcolor{lightcyan} & \methodnamethreefive       & \textbf{0.40} & \textbf{7.8} & \underline{0.010} & \textbf{0.11}\\
    \bottomrule
  \end{tabular}
  }
\end{table*}